%% file: main.tex
\documentclass[sigconf]{acmart}
\AtBeginDocument{%
  }

\usepackage{multirow}
\usepackage[ruled]{algorithm2e}
\input{macro}

\copyrightyear{2025}
\acmYear{2025}
\setcopyright{cc}
\setcctype{by}
\acmConference[CIKM '25]{Proceedings of the 34th ACM International
Conference on Information and Knowledge Management}{November 10--14,
2025}{Seoul, Republic of Korea}
\acmBooktitle{Proceedings of the 34th ACM International Conference on
Information and Knowledge Management (CIKM '25), November 10--14, 2025,
Seoul, Republic of Korea}\acmDOI{10.1145/3746252.3760801}
\acmISBN{979-8-4007-2040-6/2025/11}
\begin{document}

%%
%% The "title" command has an optional parameter,
%% allowing the author to define a "short title" to be used in page headers.
\title{In-context Pre-trained Time-Series Foundation Models adapt to Unseen Tasks}

%%
%% The "author" command and its associated commands are used to define
%% the authors and their affiliations.
%% Of note is the shared affiliation of the first two authors, and the
%% "authornote" and "authornotemark" commands
%% used to denote shared contribution to the research.
\author{Shangqing Xu}
\email{sxu452@gatech.edu}
\orcid{0009-0005-5619-1381}
\affiliation{%
  \institution{Georgia Institute of Technology}
  \city{Atlanta}
  \state{GA}
  \country{USA}
}
\author{Harshavardhan Kamarthi}
\email{hkamarthi3@gatech.edu}
\affiliation{%
  \institution{Georgia Institute of Technology}
  \city{Atlanta}
  \state{GA}
  \country{USA}
}
\author{Haoxin Liu}
\email{hliu763@gatech.edu}
\affiliation{%
  \institution{Georgia Institute of Technology}
  \city{Atlanta}
  \state{GA}
  \country{USA}
}
\author{B. Aditya Prakash}
\email{badityap@cc.gatech.edu}
\affiliation{%
  \institution{Georgia Institute of Technology}
  \city{Atlanta}
  \state{GA}
  \country{USA}
}
%%
%% By default, the full list of authors will be used in the page
%% headers. Often, this list is too long, and will overlap
%% other information printed in the page headers. This command allows
%% the author to define a more concise list
%% of authors' names for this purpose.
% \renewcommand{\shortauthors}{Xu et al.}
\renewcommand{\shortauthors}{Shangqing Xu, Harshavardhan Kamarthi, Haoxin Liu, and B. Aditya Prakash}

%%
%% The abstract is a short summary of the work to be presented in the
%% article.

\input{sections/abstract}
%%
%% The code below is generated by the tool at http://dl.acm.org/ccs.cfm.
%% Please copy and paste the code instead of the example below.
%%


% \ccsdesc[500]{Do Not Use This Code~Generate the Correct Terms for Your Paper}
% \ccsdesc[300]{Do Not Use This Code~Generate the Correct Terms for Your Paper}
% \ccsdesc{Do Not Use This Code~Generate the Correct Terms for Your Paper}
% \ccsdesc[100]{Do Not Use This Code~Generate the Correct Terms for Your Paper}

%%
%% Keywords. The author(s) should pick words that accurately describe
%% the work being presented. Separate the keywords with commas.
%% A "teaser" image appears between the author and affiliation
%% information and the body of the document, and typically spans the
%% page.

% \received{20 February 2007}
% \received[revised]{12 March 2009}
% \received[accepted]{5 June 2009}

%%
%% This command processes the author and affiliation and title
%% information and builds the first part of the formatted document.
\maketitle

\input{sections/introduction}
\input{sections/related}
\input{sections/method}

\input{sections/experiment}

\input{sections/conclusion}

\section{Acknowledgment}

This paper was supported in part by the NSF (Expeditions CCF1918770, CAREER IIS-2028586, Medium IIS-1955883, Medium IIS2106961, Medium IIS-2403240, PIPP CCF-2200269), CDC MInD program, Meta and Dolby faculty gifts, and funds/computing resources from Georgia Tech and GTRI.

\section{GenAI Usage Disclosure}

During this research, GenAI is not used for content creation, including code, data, and the written content.

\bibliographystyle{ACM-Reference-Format}
\balance
\bibliography{main}

\end{document}

%% file: macro.tex
\iftrue
    \newcommand{\hk}[1]{\textcolor{orange}{[Harsha: #1 ]}}
    \newcommand{\xsq}[1]{\textcolor{purple}{[Shangqing: #1]}}
\newcommand{\bap}[1]{{\bf Aditya says: #1}}
\else
    \newcommand{\hk}[1]{}
    \newcommand{\xsq}[1]{}
\newcommand{\bap}[1]{}
\fi

\newcommand{\model}{\textsc{ICTP}\xspace}

%% file: sections/abstract.tex
\begin{abstract}
    Time-series foundation models (TSFMs) have demonstrated strong generalization capabilities across diverse datasets and tasks. However, existing foundation models are typically pre-trained to enhance performance on specific tasks and often struggle to generalize to unseen tasks without fine-tuning. To address this limitation, we propose augmenting TSFMs with In-Context Learning (ICL) capabilities, enabling them to perform test-time inference by dynamically adapting to input-output relationships provided within the context. Our framework, In-Context Time-series Pre-training (ICTP), restructures the original pre-training data to equip the backbone TSFM with ICL capabilities, enabling adaptation to unseen tasks. Experiments demonstrate that ICT improves the performance of state-of-the-art TSFMs by approximately $11.4\%$ on unseen tasks without requiring fine-tuning.
\end{abstract}

%% file: sections/introduction.tex
\section{Introduction}

\begin{figure*}[th]
    \centering
    \includegraphics[width=0.87\linewidth]{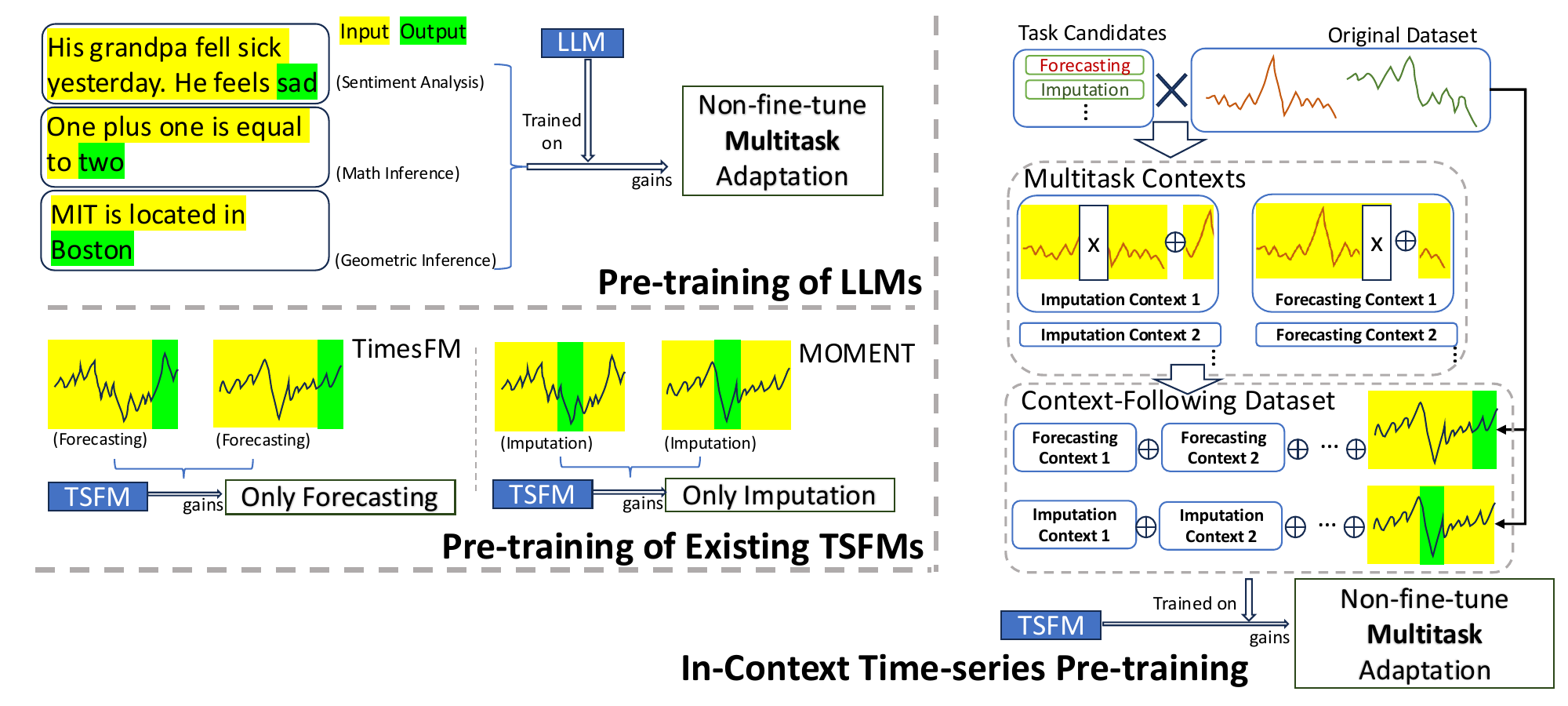}
    \caption{Similar to LLMs, existing TSFMs are pre-trained on single-task objectives using large-scale data. Yet, due to the inherent differences between time-series and language data, TSFMs are restricted to single-task adaptation. Instead, we propose In-Context Time-series Pre-training (\model), reformatting the original dataset into a multi-task context-following structure, enabling TSFMs to gain the in-context learning ability which finally leads to non-fine-tuning adaptation to unseen tasks.}
    \label{fig:ICL-timeseries}
\end{figure*}

% \bap{needs some citations}
Time-series analysis serves as a critical tool in a wide range of real-world applications, including demand forecasting \cite{suganthi2012energy, wu2021autoformer, liu2024time, zhao2025timerecipe}, pandemic analysis \cite{doornik2021modeling, rodriguez2021deepcovid}, and imputation of missing variables or historical records \cite{stefanakos2001unified, ma2019end, wu2022timesnet}. Given this diversity, the development of robust and accurate time-series models is essential—a challenge recently addressed by the emergence of time-series foundation models (TSFMs) \cite{das2023decoder, woo2024unified, goswami2024moment, kamarthi2024large}. These models are pretrained on vast datasets, enabling them to perform effectively across various tasks.

However, a key limitation of existing TSFMs is their lack of multi-task adaptation capability without fine-tuning. Current TSFMs either specialize in a single task, such as forecasting \cite{das2023decoder, woo2024unified}, or require task-specific fine-tuning before deployment \cite{goswami2024moment, kamarthi2024large, gao2024units}. This limitation increases computational overhead and data requirements, hindering their real-world applicability.

To address this gap, we propose enhancing foundation time-series models with in-context learning (ICL) \cite{brown2020language}, enabling multi-task adaptation without fine-tuning. ICL operates through a test-time inference procedure: by appending input-output pairs (the context) to the original input, the model infers the task’s requirements and produces the desired output. 

However, integrating ICL into TSFMs presents unique challenges. In language models where ICL was first realized, ICL emerges naturally from pretraining on diverse tasks embedded in textual data \cite{brown2020language, gu2023pretraining}. In contrast, time-series data lacks inherent task diversity, as datasets are uniformly structured in chronological order. Therefore, training on a single objective (e.g., forecasting) inherently restricts exposure to other tasks, making ICL acquisition impossible without explicit multi-task pretraining. To the best of our knowledge, this challenge is not tackled by the pre-training pipelines of existing TSFMs.

To overcome this, we introduce In-Context Time-series Pre-training (\model), a novel pipeline that transforms existing datasets into a multi-task format for ICL-enabled pretraining. \model first identifies task candidates and generates input-output examples for each task in a unified format, covering major sequence-to-sequence applications. Next, it constructs context sequences by combining examples from different tasks. Finally, it trains the model on these augmented sequences, explicitly teaching in-context reasoning. This approach generalizes to most time-series tasks, offering broad applicability.

To validate our approach, we train foundation models on datasets processed with \model and evaluate their performance on both seen and unseen tasks. Experimental results demonstrate that models pretrained with \model on a subset of tasks achieve significant improvements on unseen tasks. Moreover, when pretrained with \model on all tasks, the models exhibit further performance gains. We also conduct extensive ablation studies to analyze the mechanisms through which \model enhances model capabilities, providing deeper insights into its effectiveness. Our codes are available in \url{https://github.com/SigmaTsing/In_Context_Timeseries_Pretraining}.

%% file: sections/related.tex
\section{Related Work}

\subsection{Time Series Foundation Models}

% Describe current time-series foundation models, their pre-training methods, and their capability of non-fine-tune generalization (for example, Timesfm -- forecasting, Moment -- imputation)

Early TSFMs relied on repurposing~\cite{gruver2024large} or reprogramming~\cite{jin2023time,zhou2023one} existing large-language-models. 
Subsequent models are pretrained on large-scale time-series data upon objectives such as patch-level autoregressive forecasting~\cite{das2023decoder}, mask-then-reconstruction ~\cite{woo2024unified}, or maximizing likelihood of tokens ~\cite{ansari2024chronos}. However, these models are designed for a single task and exhibit limited generalization to other tasks. Meanwhile, a few recent efforts have built multi-task TSFMs by pre-training an encoder via masked reconstruction ~\cite{goswami2024moment} or adaptive segmentation~\cite{kamarthi2024large}. Nevertheless, these models require task-specific fine-tuning of projection heads before adaptation. To our knowledge, no existing TSFM achieves task-agnostic adaptation without fine-tuning.

%\subsection{In-Context Learning}

% Describe in NLP how in-context learning was introduced (by GPT3) and the effort of enhancing it (papers around enhancing icl ability like PICL, papers around example selection)
\subsection{In-Context Learning}
In-context learning (ICL), first observed in LLMs \cite{brown2020language}, enables task adaptation through dynamic prompting rather than parameter updates. While early models acquired ICL capabilities unintentionally \cite{brown2020language, chowdhery2023palm, achiam2023gpt}, recent studies proposed to actively enhance ICL through improved pre-training strategies \cite{reid2024probing, gu2023pretraining} or optimized example selection \cite{nvidia2024bayesian, xu2024misconfidence, bhope2025optiseq}. Though these advances significantly boost ICL performance in language models, their applicability to time-series domains remains unexplored.
% \citeauthor{xie2021bayesian} suggests to use ICL functions as implicit Bayesian inference.  \citeauthor{reid2024probing} further analyzes decision boundaries formed during ICL in binary classification tasks, identifying irregularities that highlight the need for improved inductive biases.\citeauthor{gu2023pretraining} explicitly trains models on intrinsic tasks to improve generalization. \citeauthor{nvidia2024bayesian} use Bayesian methods to prioritize examples based on inverse inference probabilities. \citeauthor{bhope2025optiseq} uses beam search to optimize the ordering of examples in prompts, achieving significant accuracy improvements across multiple benchmarks. \citeauthor{xu2024misconfidence} propose to iteratively refine in-context examples by estimating misconfidence. 

% Moreover, many-shot ICL has been a key research topic. \citeauthor{agarwal2024manyshot} demonstrate significant performance gains with hundreds or thousands of examples.  

%% file: sections/method.tex
% \begin{figure}[th]
%     \centering
%     \includegraphics[width=\linewidth]{figures/ICLTS2.png}
%     \caption{As natural language has various types of content that correspond to different tasks, models trained on language by a single task can observe input-output relationships from multiple tasks. Yet, time-series are strictly arranged by time. Therefore, single-task pre-trained time-series models only see one type of input-output relationship.}
%     \label{fig:ICL TS vs LLM}
% \end{figure}

\section{Methodology}

\subsection{Problem Definition} \label{sec:methodTasks}

We consider a multi-task adaptation scenario: Given a time-series $x \in \mathbb{R}^{T \times m}$, for each task $k$ from a task set $k \in K$, there's a relationship $f_k(x) = y, \ x\in D,\ y \in \mathbb{R}^{T\times h}$ between input $x$ and the desired output $y$.
% where $T$ and $h$ are the input and output horizons respectively, and $m$ is the number of channels in each vector. 
Our goal is to achieve \textit{non-fine-tune multi-task adaptation} on \textit{unseen tasks}. Specifically, a time-series model $f_\theta$, once pre-trained on a candidate task set $K = \{k_1, \dots, k_N\}$, should be able to output $f_{k}(x), k \notin K$ without fine-tuning $\theta$.

% Yet, existing models only adapt themselves to one certain task and claim that they have the non-fine-tune capability in the corresponding task. Although some models claim to have generalized embedding ability across all types of sequence-to-sequence tasks, they still need to be fine-tuned to adapt to each of them. Therefore, it's necessary to build a one-for-all model that could be used for all these three types of tasks.

% Instead of building a brand new model from scratch, we propose to enhance existing models and turn them into multi-task adapters that do not require fine-tuning. We achieve this by introducing In-Context Learning (ICL) capability.

\subsection{Enabling Non-fine-tune Multi-task Adaptation for TSFMs}

The primary challenge in achieving non-fine-tune multi-task adaptation for foundation models lies in obtaining the input-output relationship $f_k$ for various tasks without fine-tuning the model parameters. Inspired by recent progress of test-time inference in natural language processing \cite{brown2020language, gu2023pretraining} and computer vision \cite{zhou2024visual}, we address this challenge by equipping foundation models with in-context learning (ICL) capabilities.

ICL is an emergent capability first observed in large language models (LLMs) after pre-training on extensive textual corpora \cite{brown2020language}. An ICL model learns to adapt dynamically by leveraging contextual information during inference. Specifically, when an input $x$ is concatenated with a context sequence $c_k$ that contains information of a task $k$, the ICL model $f_{ICL}$ imitates $c_k$ to produce task-adapted output $f(x; c_k) \rightarrow f_k(x)$, where $c_k$ is mostly constructed by concatenating input-output pairs of task $k$. %, i.e., $c_k = \bigoplus\{x_1, f_k(x_1), x_2, f_k(x_2), \dots\}$, where $\bigoplus$ means a concatenation template.

However, extending ICL to TSFMs presents unique challenges not addressed by existing pipelines. Current TSFMs are typically pre-trained on a single-task objective (e.g., forecasting) on large-scale datasets \cite{das2023decoder, woo2024unified, goswami2024moment, kamarthi2024large}. While LLMs serendipitously developed ICL through single-task pre-training \cite{brown2020language}, subsequent research \cite{min-etal-2022-metaicl, chen-etal-2022-meta, gu2023pretraining} revealed that multitask pre-training—enabled by the inherent diversity of linguistic data—is crucial for acquiring robust ICL capabilities. For instance, next-token prediction on the sentence "The temperature dropped below zero today for the first time, so everyone is nervous" implicitly requires the model to perform sentiment analysis, demonstrating the multitask nature of language data. In contrast, time-series data follows a strict chronological order, meaning a model trained for next-step prediction will inherently specialize in short-term forecasting but fail at tasks like imputation or anomaly detection.

To overcome this limitation, we argue that time-series foundation models must be explicitly pre-trained on multiple tasks to acquire ICL capabilities. We achieve this through In-Context Time-series Pre-training (\model), a novel pipeline designed to foster multitask adaptability.

\subsection{In-Context Time-series Pre-training (\model)}

\begin{algorithm}[th]
  \caption{In-Context Time-series Pre-training (\model)}\label{alg:ictf}
  \KwData{Time-series dataset $\mathcal{D} = \{x\}$, Task candidates $K$,\\ Demonstration size $m$, TSFM $f$, \\}
  % \KwResult{Reorganized dataset $\mathcal{D}_{ICL}$, model with ICL capability $f_{ICL}$}
  $\mathcal{D}_{ICL} = \phi$\;
  \For{$x \in \mathcal{D}$}{
    Sample $k \sim K$\;
    $y^k = f_k(x)$\;
    $C_x = \phi$\;
    \For{$i$ \KwTo $m$}{
        Sample $x_i \sim \mathcal{D},\ x_i \cap x = \phi$\;
        $y^k_i = f_k(x_i)$\;
        $C_x = C_x\oplus x_i \oplus y^k_i$\;
    }
    $\mathcal{D}_{ICL} \leftarrow (C_x\oplus x, y^k)$
  }
  $f_{ICL} \leftarrow$ Finetune $f$ by $\mathcal{D}_{ICL}$
\end{algorithm}

We propose In-Context Time-series Pre-training (\model), a novel pipeline that transforms a raw time-series dataset into a multi-task context-following dataset. Given a dataset and pre-train task candidates, \model constructs input-output pairs corresponding to each data point and each task. Next, \model assembles context pieces by concatenating pairs from the same task. These context pieces are then augmented with the original inputs to form modified inputs, while the corresponding outputs remain unchanged. A complete pipeline is described in Fig \ref{alg:ictf}. We argue that pre-training TSFMs on datasets structured by \model inherently equips them with ICL capabilities, enabling multi-task adaptation without fine-tuning.

%% file: sections/experiment.tex
\section{Experiments}

\begin{table*}[ht]
\resizebox{\linewidth}{!}{
\begin{tabular}{l|l|c|cc|cc|cc|cc|cc|cc}
\toprule
\multirow{2}{*}{Backbone} & \multirow{2}{*}{Evaluated on} & \multirow{2}{*}{\model} & \multicolumn{2}{c|}{ETTh1} & \multicolumn{2}{c|}{ETTm1} & \multicolumn{2}{c|}{Exchange} & \multicolumn{2}{c|}{Weather} & \multicolumn{2}{c|}{PEMS-Bay} & \multicolumn{2}{c}{METR-LA} \\
 &  &  & MSE & MAE & MSE & MAE & MSE & MAE & MSE & MAE & MSE & MAE & MSE & MAE \\ \midrule
\multirow{4}{*}{MOMENT} & \multirow{2}{*}{Forecasting} & No & 0.813 & 0.629 & 0.724 & 0.588 & 0.228 & 0.285 & 0.215 & 0.345 & 2.623 & 0.899 & 1.291 & 0.765 \\
 &  & Yes & \textbf{0.433} & \textbf{0.458} & \textbf{0.496} & \textbf{0.541} & 0.236 & 0.279 & \textbf{0.163} & \textbf{0.206} & \textbf{1.625} & \textbf{0.595} & \textbf{1.169} & \textbf{0.728} \\ \cline{2-15} 
 & \multirow{2}{*}{BackTracing} & No & 0.834 & 0.643 & 0.75 & 0.595 & 0.229 & 0.289 & 0.222 & 0.354 & 2.594 & 0.915 & 1.284 & 0.767 \\
 &  & Yes & \textbf{0.439} & \textbf{0.454} & \textbf{0.502} & \textbf{0.527} & 0.241 & 0.305 & \textbf{0.165} & \textbf{0.254} & \textbf{1.773} & \textbf{0.613} & 1.315 & 0.780 \\ \midrule
\multirow{4}{*}{TimesFM} & \multirow{2}{*}{BackTracing} & No & 0.518 & 0.464 & 0.402 & 0.427 & 0.118 & 0.238 & 0.182 & 0.207 & 2.993 & 0.879 & 1.477 & 0.742 \\
 &  & Yes & \textbf{0.438} & \textbf{0.429} & \textbf{0.382} & \textbf{0.447} & \textbf{0.097} & \textbf{0.218} & \textbf{0.175} & \textbf{0.198} & \textbf{2.167} & \textbf{0.719} & \textbf{1.283} & \textbf{0.682} \\ \cline{2-15} 
 & \multirow{2}{*}{Imputation} & No & 0.920 & 0.604 & 0.967 & 0.649 & 0.118 & 0.244 & 0.235 & 0.281 & 2.888 & 0.835 & 1.198 & 0.613 \\
 &  & Yes & \textbf{0.785} & \textbf{0.592} & \textbf{0.934} & \textbf{0.692} & 0.134 & 0.265 & \textbf{0.231} & \textbf{0.279} & \textbf{2.818} & \textbf{0.761} & 1.412 & 0.723 \\ \midrule
\multirow{4}{*}{LPTM} & \multirow{2}{*}{BackTracing} & No & 0.830 & 0.663 & 0.739 & 0.640 & 2.259 & 1.229 & 0.471 & 0.497 & 2.832 & 0.950 & 1.528 & 0.848 \\
 &  & Yes & \textbf{0.672} & \textbf{0.597} & \textbf{0.628} & \textbf{0.564} & \textbf{2.173} & \textbf{1.134} & \textbf{0.381} & \textbf{0.435} & \textbf{2.033} & \textbf{0.688} & \textbf{1.375} & \textbf{0.733} \\ \cline{2-15} 
 & \multirow{2}{*}{Imputation} & No & 1.164 & 0.784 & 1.128 & 0.776 & 1.923 & 1.119 & 0.383 & 0.451 & 2.226 & 0.806 & 1.175 & 0.729 \\
 &  & Yes & \textbf{1.141} & \textbf{0.779} & \textbf{1.096} & \textbf{0.764} & \textbf{1.873} & \textbf{1.035} & \textbf{0.331} & \textbf{0.411} & \textbf{2.122} & \textbf{0.804} & \textbf{1.097} & \textbf{0.699} \\ \bottomrule
\end{tabular}
}
\caption{\model significantly improved the performance of backbone TSFMs on unseen tasks of output length 96, on all datasets. %Results outlined by bold shows that \model improved the performance of backbone TSFMs on unseen tasks.
} \label{tab:Results96}
\vspace{-0.25cm}
\end{table*}

\begin{table*}[ht]
\resizebox{\linewidth}{!}{
\begin{tabular}{l|l|c|cc|cc|cc|cc|cc|cc}
\toprule
\multirow{2}{*}{Backbone} & \multirow{2}{*}{Evaluated on} & \multirow{2}{*}{\model} & \multicolumn{2}{c|}{ETTh1} & \multicolumn{2}{c|}{ETTm1} & \multicolumn{2}{c|}{Exchange} & \multicolumn{2}{c|}{Weather} & \multicolumn{2}{c|}{PEMS-Bay} & \multicolumn{2}{c}{METR-LA} \\
 &  &  & MSE & MAE & MSE & MAE & MSE & MAE & MSE & MAE & MSE & MAE & MSE & MAE \\ \midrule
\multirow{4}{*}{MOMENT} & \multirow{2}{*}{Forecasting} & No & 0.938 & 0.691 & 0.849 & 0.644 & 0.337 & 0.365 & 0.506 & 0.547 & 2.247 & 0.821 & 1.336 & 0.780 \\
 &  & Yes & \textbf{0.514} & \textbf{0.527} & \textbf{0.618} & \textbf{0.544} & \textbf{0.332} & \textbf{0.345} & \textbf{0.357} & \textbf{0.427} & \textbf{2.198} & \textbf{0.698} & \textbf{1.310} & \textbf{0.760} \\ \cline{2-15} 
 & \multirow{2}{*}{BackTracing} & No & 0.944 & 0.704 & 0.886 & 0.654 & 0.337 & 0.358 & 0.556 & 0.581 & 2.296 & 0.861 & 1.319 & 0.797 \\
 &  & Yes & \textbf{0.526} & \textbf{0.512} & \textbf{0.603} & \textbf{0.531} & \textbf{0.330} & \textbf{0.347} & \textbf{0.402} & \textbf{0.495} & \textbf{2.593} & \textbf{0.847} & \textbf{1.211} & \textbf{0.728} \\ \midrule
\multirow{4}{*}{TimesFM} & \multirow{2}{*}{BackTracing} & No & 0.584 & 0.509 & 0.496 & 0.467 & 0.278 & 0.361 & 0.238 & 0.277 & 3.315 & 0.966 & 1.755 & 0.836 \\
 &  & Yes & \textbf{0.512} & \textbf{0.492} & \textbf{0.415} & \textbf{0.453} & \textbf{0.181} & \textbf{0.305} & \textbf{0.231} & \textbf{0.269} & \textbf{2.249} & \textbf{0.705} & \textbf{1.379} & \textbf{0.659} \\ \cline{2-15} 
 & \multirow{2}{*}{Imputation} & No & 1.053 & 0.642 & 0.919 & 0.610 & 0.226 & 0.334 & 0.385 & 0.382 & 3.485 & 0.986 & 1.623 & 0.752 \\
 &  & Yes & \textbf{0.913} & \textbf{0.679} & \textbf{0.852} & \textbf{0.572} & \textbf{0.242} & 0.338 & \textbf{0.374} & \textbf{0.379} & \textbf{2.570} & \textbf{0.822} & \textbf{1.516} & \textbf{0.767} \\ \midrule
\multirow{4}{*}{LPTM} & \multirow{2}{*}{BackTracing} & No & 0.811 & 0.859 & 0.765 & 0.658 & 2.456 & 1.314 & 0.435 & 0.468 & 2.841 & 0.965 & 1.481 & 0.822 \\
 &  & Yes & \textbf{0.796} & \textbf{0.661} & \textbf{0.627} & \textbf{0.568} & 2.645 & 1.359 & \textbf{0.431} & \textbf{0.473} & \textbf{1.859} & \textbf{0.605} & \textbf{1.375} & \textbf{0.807} \\ \cline{2-15} 
 & \multirow{2}{*}{Imputation} & No & 1.244 & 0.818 & 1.403 & 0.861 & 2.016 & 1.156 & 0.515 & 0.497 & 2.618 & 0.909 & 1.330 & 0.774 \\
 &  & Yes & \textbf{1.164} & \textbf{0.785} & \textbf{1.168} & \textbf{0.784} & 2.199 & \textbf{1.146} & \textbf{0.343} & \textbf{0.413} & \textbf{2.508} & \textbf{0.889} & \textbf{1.294} & \textbf{0.787} \\ \bottomrule
\end{tabular}
}
\caption{\model significantly improved the performance of backbone TSFMs on unseen tasks of output length 192, on all datasets.
%Results outlined by bold shows that \model improved the performance of backbone TSFMs on unseen tasks.
}\label{tab:Results192}
\vspace{-0.25cm}
\end{table*}

\subsection{Settings} 

% To demonstrate the effect of \model on TSFMs' performance of unseen tasks, we collect several time-series task candidates, applying \model to pre-train backbone TSFMs while iteratively excluding one of the task candidates, and compare the pre-trained TSFMs on the excluded task without fine-tuning to see if their performance on such an unseen task has been improved.

\subsubsection{Backbone Models}

We choose three representative foundational time-series models—MOMENT \cite{goswami2024moment}, TimesFM \cite{das2023decoder}, LPTM \cite{kamarthi2024large}— as the backbones. Such a selection covers three pre-training objectives commonly adapted in time-series models: mask construction, autoregressive generation, and adaptive segmentation.  % While incorporating \model, we keep the original pre-training objective for each model, only reforming the pre-training dataset using \model.

\subsubsection{Tasks Candidates}

We collect three time-series tasks as candidates for pre-training and evaluation: 1) Forecasting, where the model predict the consecutive future values of a given sequence; 2) Imputation, where the model rebuilds certain part of the input which was masked 3) Backtracing, where the model predict the consecutive history values of a given sequence. While evaluating \model on each task, we pretrain backbone models on the other tasks, keeping the evaluation task unseen. For example, while evaluating the models on backtracing, the task candidates in \model will be forecasting and imputation. Specifically, as TimesFM and LPTM have accessed forecasting data during their original pre-training procedure and MOMENT has accessed imputation data, we exclude these tasks in corresponding evaluation. 

\subsubsection{Baselines}

As the scope of \model is to adapt single-task foundation models to multiple tasks without fine-tuning, we set baseline methods as task-aware naive input reprogramming that does not require fine-tuning. 

For TimesFM and LPTM, we 1) truncates imputation inputs before (after) the imputation target, depending on whether the reconstruction area surpasses the middle of the original sequence, while maintaining chronological order 2) flip the backtracing inputs and outputs.

For MOMENT, we 1) concatenate forecasting inputs and outputs as input, with mask on original outputs 3) flipped the backtracing inputs and outputs, then apply the same strategy as forecasting.

% We don't include forecasting for decoder-only models and imputation for encoder-only models in comparison because such models are already pre-trained on the corresponding tasks. Still, we show in Sec \ref{sec:ablation} how \model affects the zero-shot performance of original tasks on the corresponding models.

\subsubsection{Dataset}
We choose four datasets (ETTh1, ETTm1, Exchange Rate, Weather) from the Informer datasets \cite{zhou2021informer}, and two datasets (PEMS-Bays, and METR-LA) from DCRNN datasets \cite{li2017diffusion} to pre-train and evaluate \model. For all the datasets, we adopt Channel Independence assumption \cite{nie2022time}, conducting tasks on each channel only considering inputs from the corresponding channel. We split the train / valid / test data chronologically by 60:20:20 (that is, train data is always earlier than valid / test). We referred to implementations of TimesNet \cite{wu2022timesnet} to normalize the input data. 

For all tasks, we consider two output lengths, 96 and 192, for all datasets. For forecasting and backtracing, the lookback window is set as 192 and 384, respectively. For Imputation, we set the input length as 192 and 384 correspondingly and randomly mask 96 (192) of the inputs as reconstruction targets so that the input/output length all aligns between different tasks. We use 4 context examples in all tasks. While building context sequences, we make sure there's no overlap between examples and target output. 

% Although many of the foundation models have proposed to use sequences as long as possible to be the input, we claim that such an approach greatly relies on the capability of backbone models and, therefore, cannot be generalized. 

% As for adopting \model, we propose that, instead of building a standalone pre-training pipeline and training a brand new model from scratch, it's more efficient to apply \model on existing checkpoints of foundation models and treat \model as a second-stage enhancement. Therefore, we adapt \model on foundation models that are already pre-trained on certain tasks, expecting \model to bring enhancement to their unseen task performance.

% We calculate two metrics for each experiment: Mean Squared Error (MSE) and Mean Absolute Error (MAE). Given an output sequence $y' = \{y'_1, \dots, y'_M\}$ and a ground truth $y = \{y_1, \dots, y_M\}$, these two metrics are calculated by:

% \begin{equation*}
%     MSE(y, y') = \frac{1}{M} \sum_j (y_j - y_j')^2 \ \ \ MAE(y, y') = \frac{1}{M} \sum_j |y_j - y_j'| 
% \end{equation*}

\subsection{Results}

% We set up two experiments to ablate the \model's effect. In the first setting, we include all tasks during \model and evaluate the model on each task. This setting aims to show the joint effect of obtaining multiple tasks other than doing a single task pre-training. In the second setting, we exclude the target task from \model and only pre-train on the other tasks, making sure the evaluation tasks are unseen during the pre-train procedure. This will give us insight into how \model helps on unseen task performance, which is our goal.

The results are presented in Table \ref{tab:Results96} and \ref{tab:Results192} for output lengths of 96 and 192, respectively. After adapting \model, the backbone TSFMs exhibited significant performance improvements on unseen tasks across most datasets, with average improvement ratios of $11.3\%$ and $11.6\%$, respectively. This demonstrates that \model effectively enhances the capability of TSFMs to handle unseen tasks. Notably, this improvement is achieved without any prior knowledge of the downstream task, highlighting \model's potential for broader applications.

It's worthy noticing that the degree of improvement varies across models and datasets. Specifically, \model struggles to enhance imputation performance for decoder-only models (TimesFM, LPTM) but shows greater success in improving forecasting and backcasting performance for encoder-only models (MOMENT). Additionally, the improvement on the Exchange dataset is the smallest among all datasets. We attribute this to the dataset's simplicity: unlike the others, Exchange consists of weekly foreign currency exchange rates, which exhibit relatively smooth patterns. Consequently, TSFMs can more easily adapt to the input-output variance gap across tasks in this case. 

Furthermore, we note that the Weather dataset was already included in TimesFM's original pre-training process. Despite this, \model still substantially improved TimesFM's performance on unseen tasks. This supports our hypothesis that existing pre-training pipelines for TSFMs, while effective for specific tasks, do not inherently equip models with multi-task capability without fine-tuning, which is a gap that \model successfully addresses.

%% file: sections/conclusion.tex
\section{Conclusion and Discussion}
In this paper, we present In-Context Time-series Fine-tuning (\model), a novel method for enhancing the non-fine-tuning adaptability of TSFMs on unseen tasks. By restructuring pre-training datasets to incorporate multi-task coverage and explicit context paradigms, \model equips TSFMs with ICL capabilities akin to those of LLMs. Our experiments demonstrate that \model significantly improves performance on unseen tasks while maintaining robust performance on previously encountered tasks. Future work could explore extending \model to a broader range of tasks or more diverse real-world datasets. 

% This work highlights the potential of \model as an efficient and scalable approach to enable multi-task in-context learning for time-series models. Beyond these promising results, this work still has several limitations. First, the evaluation has been tested only on regular datasets and domains and doesn't consider out-of-distribution tasks \cite{liutime}. Second, the solution primarily focuses on sequence-to-sequence tasks and may require additional efforts to support other task formats, such as reasoning \cite{liu2024picture}. 